%% file: 0_main.tex
\documentclass[twocolumn]{svjour3}

\usepackage{graphicx}
\usepackage{float}
\graphicspath{{Figures}}
\DeclareGraphicsExtensions{.pdf,.png}
\usepackage{url}
\usepackage{xcolor}
\usepackage{hyperref}
\hypersetup{
     colorlinks=true,
     linkcolor=orange,
     filecolor=orange,
     citecolor=orange,      
     urlcolor=orange,
     }
\usepackage{color, colortbl}
\usepackage{amsfonts}
\usepackage{amsmath}
\usepackage{appendix}
\usepackage{algorithm, setspace}
\usepackage{algpseudocode}
\usepackage{csquotes}
\usepackage{amsfonts}
\usepackage{booktabs}
\usepackage{siunitx}
\usepackage{breqn}
\usepackage[normalem]{ulem}
\usepackage{tikz}
\usetikzlibrary{tikzmark}
\usepackage{caption}
\usepackage{tcolorbox}
\tcbuselibrary{breakable}
\usepackage{placeins}
\usepackage{listings}

\lstdefinelanguage{json}{
    basicstyle=\ttfamily,
    numbers=none,
    breaklines=true,
    showstringspaces=false,
    tabsize=4
}

\newcommand{\p}[1]{\medskip \noindent \textbf{{#1}.}}
\newcommand{\eq}[1]{Equation~(\ref{eq:#1})}
\newcommand{\fig}[1]{Figure~\ref{fig:#1}}

\journalname{Autonomous Robots}
\begin{document}

\title{
VOLT: Vision and Language Trajectory Segmentation \\for Faster-than-Demonstration Policies
}


\author{Robert Ramirez Sanchez$^{1}$ \and Daniel J. Evans$^{1}$ \and Dylan P. Losey$^{1}$ \and Siddarth Jain$^{2}$}

\institute{
Robert Ramirez Sanchez \\
\email{robertjrs@vt.edu} \\
\\
Daniel J. Evans \\
\email{danielevans@vt.edu}\\
\\
Dylan P. Losey \\
\email{losey@vt.edu} \\
\\
Siddarth Jain \\
\email{sjain@merl.com}
\\
\at $^{1}$ Mechanical Engineering Department, Virginia Tech, Blacksburg, USA
\\
\at $^{2}$ Mitsubishi Electric Research Laboratories (MERL), Cambridge, MA, USA
}

\maketitle
\begin{abstract}

Humans often take longer to demonstrate a task than a robot would need to execute it. Rather than learning to replicate the demonstration at the same pace, many industrial and practical applications require robots to perform tasks as quickly as possible. In this paper, we investigate several hypotheses for learning policies that operate \textit{faster-than-demonstrations}. Our experiments show that the most effective strategy is to downsample recorded demonstrations and train the robot's policy on this accelerated data. However, uniformly downsampling an entire trajectory can be problematic. Some parts of a task can be safely sped up (e.g., unconstrained motion), while others demand slower, more precise motion (e.g., object interactions or fine manipulation). To address this challenge, we introduce VOLT, a vision-and-language trajectory segmentation method that reasons over video demonstrations, and leverages contextual cues to determine when acceleration is appropriate and when careful precision is required. VOLT identifies segments where slow, deliberate motion is necessary, then selectively downsamples the remaining segments. The resulting reformatted trajectories can be used with standard imitation learning approaches, such as diffusion policies. Our results highlight that segmentation quality is critical---baseline methods often misidentify when acceleration is possible, leading to overly cautious or unreliable policies. Compared to state-of-the-art alternatives, VOLT allows robots to execute tasks faster while maintaining strong performance. See our project website: \url{https://volt2026.github.io/VOLT2026/}

\end{abstract}



\input{1_intro}
\input{2_related}
\input{3_problem}

\input{4_method}
\input{5_experiments}
\input{6_conclusion}
\input{7_declaration}

\bibliographystyle{spmpsci}
\bibliography{references}

\input{8_appendix}

\end{document}

%% file: 1_intro.tex
\section{Introduction}\label{sec:intro}

\begin{figure*}[t]
    \begin{center}
    \includegraphics[width=\linewidth]{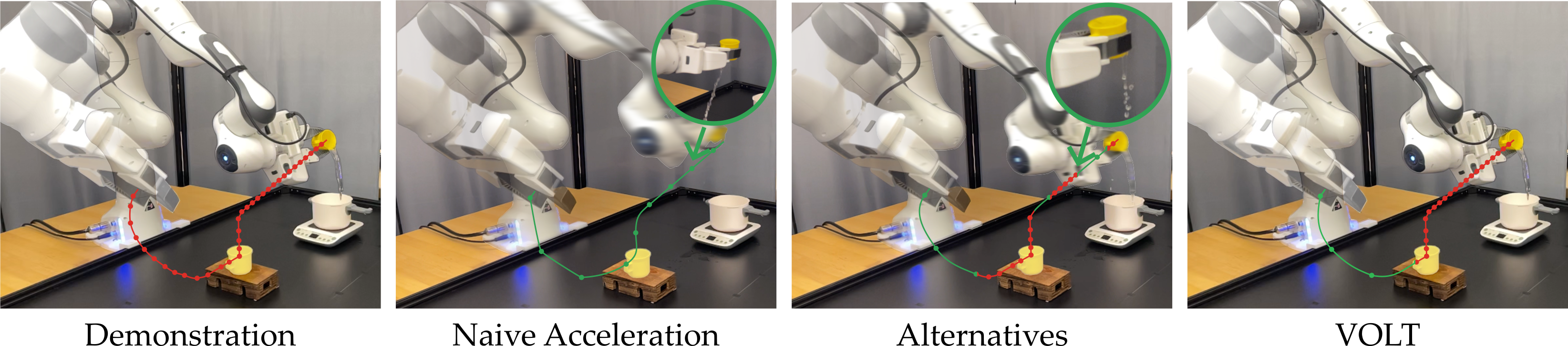}
        \caption{(Left) A human shows the robot how to carry a cup full of water and then pour that water into a pot. (Middle) Naively accelerating the entire demonstration causes the robot to move too quickly (faster motions shown in \textcolor{green}{green}), accidentally spilling water as it moves. Alternatives slow down the robot in segments (slower motions shown in \textcolor{red}{red}). But these segments are often imperfectly chosen, and the robot still accelerates and spills water while carrying the cup. (Right) VOLT addresses this problem by using context cues from the entire video demonstration to accelerate only the parts of the demonstration that do not require precision.}
        \label{fig:front_figure}
    \end{center}
\end{figure*}

Robots should be capable of performing tasks with precision and speed. Recent works have enabled robots to learn complex tasks by imitating human experts~\cite{zare2024survey,kawaharazuka2025vla,tsuji2025survey}. Since the quality of demonstrations dictates the performance of imitation learning approaches~\cite{xu2022discriminator,2023dataquality,chen2025curating}, operators tend to demonstrate tasks at a lower speed to minimize errors. 
Indeed, even if the human demonstrates at a normal speed --- robots often learn to perform tasks slower than demonstrated.
But autonomous systems are capable of completing tasks significantly faster than human demonstrators, and this acceleration is beneficial for industrial applications. 
So how can robots take a given demonstration and speed-up their learned policy?

Existing works often attempt to accelerate learned policies during either training or execution.
For example, \textit{training} approaches include high-frequency controllers that reduce teleoperation time~\cite{zhao2023learning}, as well as data augmentation methods that resample the provided demonstrations~\cite{nam2025speedaug,guo2025demospeedup}.
At the other extreme, \textit{execution} approaches accelerate a learned policy by leveraging high-fidelity controllers to precisely track the original behavior at variable frequencies~\cite{arachchige2025sail}.
Both types of approaches recognize two key problems.
First, the robot needs to decide \textit{how} to accelerate its policy, and second, the robot must determine \textit{which parts} of its trajectory are appropriate to accelerate.
Speeding up the motion often leads to failures, particularly when the robot must perform precise subtasks. (\fig{front_figure}).

In this work we seek a fundamental understanding of both questions:
(a) how do we determine which segments of a demonstration to accelerate, and 
(b) what approaches are most effective for accelerating those segments.
We adopt an experimental approach, and test a variety of hypothesis across a standardized testing setup.
Our insight --- consistent with prior works --- is that downsampling the human's demonstration provides an effective method for speeding-up the learned policy.
However, unlike prior works, we argue that high-level task knowledge is often necessary to determine which parts of the trajectory to accelerate.
Specifically, we hypothesize that:
\begin{center}
    \textit{Reasoning over videos of the entire demonstration provides the context cues needed to decide when to accelerate.}
\end{center}
We accordingly apply vision and language models (VLMs) to develop \textbf{VOLT}.
The key idea for VOLT is that the robot collects its standard visual demonstration, and then passes that demonstration through a VLM to determine the correct segmentation.
After determining which parts the robot can safely accelerate, VOLT then downsamples those segments and passes the refined dataset to the given imitation learning algorithm.
This enables the robot to learn a policy that is faster than the human's demonstration (e.g., speeding up to quickly reach the cup), but also effective at completing the task (e.g., slowing down to carefully grasp that cup). In summary, this work makes the following contributions:

\p{Accelerating Robot Policies} We introduce VOLT, a framework for enabling faster-than-demonstration policies. To the best of our knowledge, VOLT is the first approach to use a VLM to holistically analyze human demonstrations without explicitly defined features, segmenting them into regions that can be safely accelerated and regions that require precise execution.

\p{Determining How to Accelerate} We develop and test a variety of hypotheses for effective faster-than-demonstrator policies. 
These include accelerating policy execution at test-time and learning faster policies via train-time downsampling. Our results suggest that train-time downsampling is a more consistent approach, particularly when using diffusion policies.

\p{Comparing to Baselines} We compare VOLT to state-of-the-art baselines for faster-than-demonstration policies across a diverse range of tasks. Overall, we find that VOLT achieves up to a x$2.57$ speedup over policies trained on raw demonstrations while maintaining similar success rates.
We conclude with practical guidelines for designers who want to accelerate policies within an imitation learning context.

%% file: 2_related.tex
\section{Related Work} \label{sec:related}

Emerging work in imitation learning enables robots to complete complex and long-horizon tasks~\cite{black2024pi0,dai2025civil,wang2023mimicplay,zhao2025cot}. 
In general, these methods focus on matching the action distribution of the human's dataset.
Recent research suggests that robots can improve their performance and generalization abilities by learning task-relevant representations~\cite{sanchez2025recon,zhu2023learning}, leveraging language-based reasoning for task planning~\cite{clark2025action,jones24fuse}, and learning from large-scale robot datasets~\cite{kim2025openvla,bjorck2025gr00t}. 
But these methods primarily focus on task success --- in this paper, we are interested in the task speed.
Human demonstrations are typically slow to avoid mistakes, and even if the human moves quickly, for industrial applications we often want robots to operate beyond the human's capabilities. Our VOLT framework is not tied to any specific imitation learning algorithm.
But to provide results that are consistent with the state-of-the-art methods, we will leverage diffusion policies in our experiments \cite{chi2025diffusion}.

\p{Unintentional Speed Up}
Within the context of imitation learning, several methods alter the given dataset to improve data efficiency and mitigate compounding errors~\cite{parekh2026towards,shi2023waypoint,gao2024prime,belkhale2023hydra}. 
The common theme across these works is to reduce the prediction horizon required to complete a task.
Although accelerated execution is not the objective, the changes these methods make to the demonstrations often result in a sped-up policy.
For example, Belkhale \textit{et al.}~\cite{belkhale2023hydra} propose a hybrid architecture that dynamically switches between waypoints and dense low-level actions; the robot moves more rapidly between the waypoints marked by the human teacher. 
Similarly, Shi \textit{et al.}~\cite{shi2023waypoint} introduce Automatic Waypoint Extraction (AWE), a method to automatically extract waypoints from trajectories via linear interpolation within a specified error threshold. 
A different perspective is to decompose the trajectory into a set of primitives, and then learn a policy to sequence these primitives --- again, reducing the prediction horizon~\cite{gao2024prime}. 
The insight we obtain from these works is that data alterations are an effective way to reach faster-than-demonstration policies.
In VOLT we extend these ideas, while specifically focusing on how rapidly the learned policy performs the entire task.

\p{Intentional Speed Up} Recent works seek to maximize the robot's speed without significantly impacting its performance~\cite{nam2025speedaug,yuan2025speedtuning,guo2025demospeedup,arachchige2025sail,kim2025espada}. In reinforcement learning, existing works include learning speed interpolation policies~\cite{yuan2025speedtuning}, and performing online fine-tuning to select faster actions~\cite{nam2025speedaug}. Although these approaches can capture complex speed-up strategies, they typically require millions of online interactions, making them impractical in most real-world tasks.
Most similar to our approach are offline imitation learning methods that label each demonstration by classifying states suitable for faster task execution. 
Guo \textit{et al.}~\cite{guo2025demospeedup} learn a proxy policy from the original dataset to estimate the conditional action entropy distribution. 
They denote high-entropy actions as ``casual'' segments and low-entropy actions as ``precise''. 
By contrast, \cite{arachchige2025sail} estimates motion complexity via AWE~\cite{shi2023waypoint} waypoints and DBSCAN~\cite{ester1996density} clustering to identify precise segments. 
We hypothesize that these approaches can be inaccurate because they do not consider the high-level context of tasks.
Towards this objective, \cite{kim2025espada} explicitly defines $3$D gripper-object distances as high-level context extracted from images. They estimate these features using VLMs, then parse them into text for an LLM to label a single demonstration, and propagate these labels via Dynamic Time Warping. But there are human intents not encoded by these designer-chosen features, for example, gripper-object interaction (e.g., grasping vs. pushing), or manipulation of deformable bodies.In VOLT we do not assign intermediate features; instead, we directly use VLMs to reason over video demonstrations.


%% file: 3_problem.tex
\section{Problem Statement} \label{sec:problem}

We consider visual imitation learning from offline datasets of human demonstrations, where a robot arm is teleoperated by a human while the system records environment states and actions. 
Because the quality of the learned policy is correlated to demonstration fidelity, experts typically perform demonstrations slowly and cautiously to avoid mistakes. 
This results in policies that can achieve high task success but execute behaviors at relatively low speeds. 
The central problem we address is how to improve execution efficiency, specifically, increasing task execution speed without degrading task performance or safety. 
To this end, we aim to (a) autonomously identify which portions of a task can be safely executed at higher speeds, and then (b) develop a strategy for accelerating those segments without degrading overall performance.

\p{Policy} 
Let $x \in \mathcal{X}$ represent the robot joint and gripper position, and let $y \in \mathcal{Y}$ denote the environment state observed through RGB cameras. 
We combine both the robot state and RGB images into observation $o_t = (x_t, y_t)$.
The robot's learned policy $\pi$ maps these observations into actions:

\begin{equation} \label{eq:P1}
    A_t \sim \pi(\circ \mid O_t)
\end{equation}

where $A_t = \{a_{t}, a_{t+1}, ..., a_{t+H}\}$ is a sequence of actions with horizon $H$, and each action $a_t$ is a commanded robot joint and gripper position.
The observation history $O_t = \{o_{t-M}, ..., o_{t}\}$ is composed of the most recent $M$ observations. 

In our experiments we will instantiate \eq{P1} as a diffusion policy~\cite{chi2025diffusion}.
To match state-of-the-art performance, we will apply the Denoising Diffusion Implicit Models approach~\cite{song2021denoising}; this requires fewer denoising iterations during inference (as compared to training) for faster action generation.
The inference process is defined by the following equation:

\begin{equation}
\label{eq:diffusion}
A_{t}^{k-1} = \alpha(A_{t}^{k}-\gamma \varepsilon_{\theta}(O_{t},A_{t}^{k},k)+\sigma),\; \sigma\sim\mathcal{N}(0,\sigma^{2}I)
\end{equation}

where $\varepsilon_{\theta}$ is the noise prediction network with learnable parameters $\theta$ that are optimized to predict the noise added to each sample. The noise scheduler hyperparameters $\alpha$, $\gamma$, and $\sigma$ control the signal scaling and the magnitude of Gaussian noise added to each iteration $k$.

\p{Labeling Dataset}
To train $\pi$ the human provides a dataset of task demonstrations $\mathcal{D} = \{ \xi_{0}, \xi_{1}, ..., \xi_{N}\}$. 
Each demonstrated trajectory $\xi$ is a sequence $T$ of observation-action pairs: $\xi = \{ ((x_{0}, y_{0}), a_{0}), ..., ((x_{T}, y_{T}), a_{T}) \}$. 
As we will show, learning a policy directly from $\mathcal{D}$ limits the robot's speed to that of the human teacher.
One potential way to accelerate the learned policy is to first relabel each demonstration $\xi = \{(\tau_0, l_0), ..., (\tau_{k}, l_{k})\}$ into ordered non-overlapping segments $\bigcap_{i=1}^{k}\tau_i =\emptyset$.
The labels $l_i \in L = \{\textit{speed-up, maintain-speed}\}$ denote whether the given segment can be executed faster than demonstrated, or whether the segment requires precision or dexterity (e.g., grasping, insertion) and therefore the policy should match the human's original speed.

\p{Hierarchical Control} 
Actions $a$ indicate the desired joint position and gripper state.
To execute the policy quickly, the robot must select control inputs that rapidly perform these actions.
In our experiments we update the target robot position at $20$ Hz (e.g., we output $20$ actions $a_t$ per second).
We then reach these target positions using a low-level velocity controller that operates at $1$ kHz.
The specific controller depends on the robot (in our experiments, we tested with a Franka Emika robot arm).
But we generally focus on problem settings where the robot has a high-level diffusion policy --- outputting desired positions $a$ --- and a low-level controller --- which can accurately track those targets.

%% file: 4_method.tex
\section{How Do We Enable Faster-than-Demonstration Policies?} \label{sec:method}

\begin{figure*}[h!]
    \begin{center}
        \includegraphics[width=\textwidth]{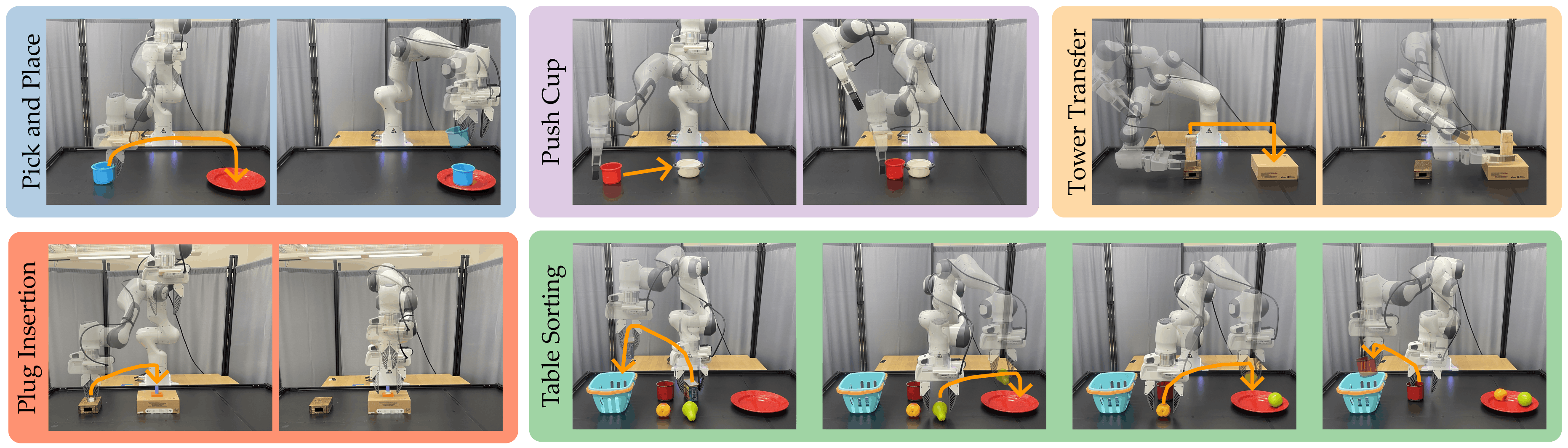}
        \caption{Real-world manipulation tasks evaluated in our experiments. The manipulated objects are randomly initialized in each task, and the target object location is fixed. (Top) We include short-horizon tasks that require object transport: \textit{Pick and Place} and \textit{Tower Transfer}, and non-prehensile manipulation in \textit{Push Cup}. (Bottom) \textit{Plug Insertion} is a challenging task requiring precise control and force-based interaction, and \textit{Table Sorting} is a long-horizon task where the robot sorts fruits on a plate and containers in a basket.}
        \label{fig:experiments_tasks}
    \end{center}
\end{figure*}

Our goal is to develop robot policies that maximize task speed while maintaining optimal performance. 
However, simply training a policy $\pi$ on collected trajectories $\mathcal{D}$ will inherently limit the robot's motion to the pacing of the human demonstrator. 
In this section, we therefore test different methods that designers might intuitively consider for speeding up the robot's actions.
We divide these methods into two groups: accelerating the policy during \textit{execution} (i.e., reducing the number of high-level actions to shorten the task), or accelerating the policy during \textit{training} (i.e., downsampling the dataset to emulate a faster demonstration).
Our experiments in Sections~\ref{sec:method_a} and \ref{sec:method_b} suggest that the training paradigm is more effective for enabling faster policies and retaining the policy's success rate. 

After we collect this experimental data for faster-than-demonstration policies, we next need to determine which parts to accelerate. We find that naively speeding up \textit{all} the robot's actions exacerbates failures, particularly when the robot requires precise motion.
We address this problem through our proposed VOLT algorithm in Section~\ref{sec:method_c}.
VOLT reasons over video demonstrations and segments these trajectories into sections that should either be accelerated or left at the demonstrated speed.
We downsample the marked segments and train the robot; our resulting policies yield faster execution while maintaining task performance. However, we note that there are practical limits on how much the robot can accelerate, even after correctly identifying these segments. 
In Section~\ref{sec:method_d} we discuss the practical factors for maximizing speed.




\p{Baseline Policy} 
We train diffusion policies using the original dataset (\textit{Normal Speed}) as our baseline for performance and task speed. 
Diffusion policies are capable of accurately learning action chunks, which are crucial for speeding up execution.  
In our experiments, inference delay is about $\sim0.1$ seconds to generate an action sequence of 32 actions. Since the inference time is greater than the desired $20$ Hz high-level control frequency, we eliminate execution delays caused by sensing-inference delays through asynchronous inference, and enforce temporal consistency via Error-Adaptive-Guidance (EAG)~\cite{arachchige2025sail}. During training, we utilize reached robot poses $x_{t+1}$ instead of the commanded poses $a_t$ as our policy actions because the low-level controller can track the former with minimal error. These additions to the standard diffusion policies were made to facilitate acceleration.

\p{Acceleration} Within this section we will test two paradigms for faster-than-demonstration policies. 
These paradigms perform acceleration during either  execution (predicted action downsampling) or during training (demonstration downsampling):

\begin{figure*}[ht]
    \begin{center}
    \includegraphics[width=\textwidth]{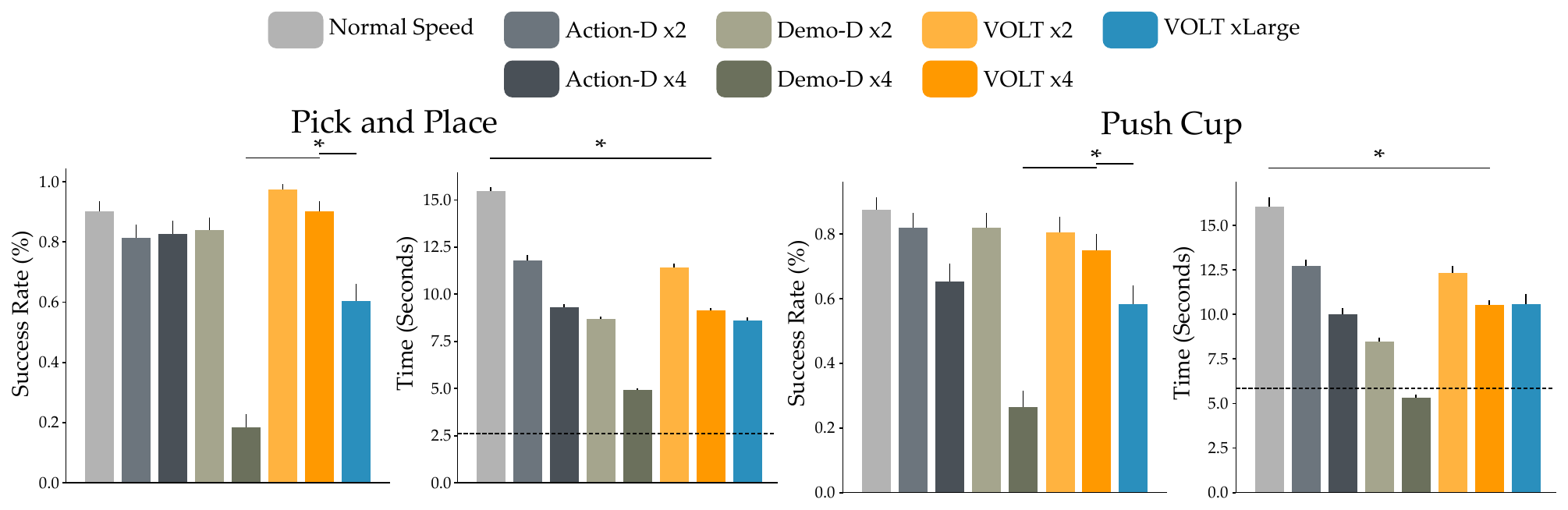}
    \caption{Experimental results for Section~\ref{sec:method}. Results are averaged across $3$ training runs. \textit{Normal Speed} is trained with the default dataset. In \textit{Action-D} action sequences are downsampled at run time to accelerate task execution. By contrast, in \textit{Demo-D} actions are downsampled at training time to emulate a human with faster teaching motions. \textit{VOLT} is similar to \textit{Demo-D}, but it uses a VLM to determine which segments of the trajectory should or should not be downsampled. The downsampling rate is captured by $n$; i.e., in $\times2$ the robot skips every other action. We find that speeding up the policy during execution (\textit{Action-D}) degrades performance as $n$ increases, since the policy misses critical actions (e.g., aligning the grippers). Similarly, speeding up the entire policy during training (\textit{Demo-D}) lowers the success rate. VOLT is able to maintain its success by only accelerating the parts of the task that do not require precise motion. However, the performance of VOLT still degrades if the human increases $n$ to a level where the low-level controller is unable to track the commanded actions (i.e., there are still practical limitations). The dotted lines in Time denote the lower bound for \textit{VOLT} if the robot could instantaneously complete the accelerated segments. An $*$ denotes statistical significance, and error bars indicate standard error.}
    \label{fig:hypotheses_test}
    \end{center}
\end{figure*}

\begin{itemize}
    \item 
    \textit{Action Downsampling (Action-D)}: uniformly downsample the predicted action chunk $A_t$ with a fixed factor $n$. Instead of performing actions $a_1, a_2, \text{etc.}$, here the robot performs actions $a_1, a_{1+n}, \text{etc.}$ This shortens the task into fewer steps and thereby accelerates the robot's execution.

    \item 
    \textit{Demonstration Downsampling (Demo-D)}: uniformly downsample each demonstration by factor $n$. Instead of training on trajectory $\xi = \{(o_1, x_1), (o_2, x_2), ...\}$, here the robot trains with $\xi = \{(o_1, x_{1+n}), (o_{1+n}, x_{1+2n}), ...\}$. Downsampling makes the demonstration shorter, resulting in a faster learned policy.
\end{itemize}

\p{Evaluation Setup}
We perform all experiments on a  Franka Emika robot arm on a tabletop setup as shown in \fig{experiments_tasks}. 
We collect all demonstrations using a GELLO~\cite{wu2024gello} to teleoperate the arm at $20$ Hz.
For each timestep, we record the robot's joint position, including the gripper state $x$, the commanded joint and gripper state $a_t$, and RGB images $y$ from three RealSense D435 cameras. 
Two of the cameras are fixed in the environment, and one is directly mounted on the robot's end-effector. 
Each paradigm is tested on the following manipulation tasks:

\begin{itemize}
    \item \textit{Pick and Place}: The robot grasps a blue cup located on the table and places it on top of a red plate.

    \item \textit{Push Cup}: The robot pushes a red cup to the nearest side of a white bowl while maintaining contact only with the outside of the cup.
\end{itemize}

\subsection{Acceleration During Execution} \label{sec:method_a}

Given a trained policy on the original dataset, the first strategy we consider is speeding up its execution. We take a policy $\pi$ and modify its action outputs so that the robot completes the task in less time. The way we modify these actions depends on the policy structure. Referring back to \eq{P1}, our diffusion policy outputs a sequence $A$ of actions, where each action is a commanded joint position and gripper state.
We can accelerate the robot's motion by \textit{skipping} actions within the sequence, i.e., downsampling $A$ by a fixed rate $n$.
After downsampling, the new sequence $A'$ has a reduced horizon; thus, the execution time of a single prediction decreases linearly as $n$ increases. 
Since the prediction horizon has fixed length $H$, the downsampling rate $n$ is constrained to at most executing one action ($n \leq H - 1$). 
Note that omitting parts of the predicted actions makes the scheduled trajectory lose fine-grained details, and the low-level control has to apply larger control signals to track the desired states in $A'$, due to the increased distance in consecutive actions. 
Considering that completing a task may not require the robot to visit all states that the human demonstrated, this uniform acceleration could potentially speed up task execution.

\fig{hypotheses_test} compares the performance of \textit{Normal Speed} against test-time downsampling (\textit{Action-D}) at rates $n=2$ and $n=4$. As expected, the total time decreases as $n$ increases. However, the performance of the learned policy degrades, particularly in tasks where the robot needs consistent, fine-grained motions. We observed that uniformly downsampling predicted action chunks often prevented the robot from executing actions critical for task completion. For example, when moving to push the cup, the robot needed to align its gripper behind the cup before initiating the push motion; if the accelerated actions $A'$ skip these steps, then the robot failed the task. A more fundamental issue, however, was the mismatch between the robot's speed during training and testing. The observation history $O_t = \{o_{t-M}, ..., o_{t}\}$ used to train $\pi$ was obtained at the human's teaching speed. When we accelerate during execution, the robot collects observations $O$ with larger motions between images, pushing the robot's policy farther out-of-distribution as $n$ increases. This mismatch between behaviors the policy has been trained upon and the behaviors expected at test time limits acceleration during execution and degrades task performance.

\subsection{Accelerating During Training} \label{sec:method_b}

With these issues in mind, we next explore methods that downsample the human's offline dataset.
The core idea here is that --- by reducing the number of datapoints --- we effectively make the human's demonstration faster.
Policies trained on the accelerated dataset expect rapid movement during execution, and thus do not have the out-of-distribution challenge faced in Section~\ref{sec:method_a}.
In practice, we perform downsampling by dividing a single trajectory into $n$ non-overlapping versions.
The first version is $\xi^1 = \{(o_1, a_1), (o_{1+n}, a_{1+n}), \ldots\}$, the second version is $\xi^2 = \{(o_2, a_2), (o_{2+n}, a_{2+n}), \ldots\}$, and so on.
Similar to before, we train new policies with $n=2$ and $n=4$, and test their performance in \fig{hypotheses_test}.

We find that \textit{Demo-D} has the potential to significantly reduce execution time, and performs the task faster than equivalent versions of \textit{Action-D}.
But this speed comes at a cost: increasing to $n=4$ sharply reduces the success rate, significantly under-performing the alternatives.
Observing the robot's motion, we find that \textit{Demo-D} fails because the robot is trying to complete segments of the task \textit{too rapidly}.
For example, in the pick and place task the robot often misses the cup, and in the pushing task the robot often knocks the cup instead of sliding it across the table.
The policy tries to address these mistakes --- but by the time the robot moves to fix errors, the end-effector is already far from the cup and out-of-distribution.
Accelerating during training has potential, but naively downsampling the entire motion results in critical failures.

\subsection{VOLT: Vision and Language Trajectory Segmentation} \label{sec:method_c}

Now that we have a method for accelerating the robot's policy, we will determine \textit{when} to speed up the robot's motion.
Our intuition is that the videos included within the demonstration provide the context necessary for segmenting the trajectory.
For instance, through these videos we can separate between a robot moving close to a cup unintentionally, and a robot moving towards a cup to grasp it.
Based on this insight, we propose \textbf{VOLT}: \textbf{V}isi\textbf{o}n and \textbf{L}anguage \textbf{T}rajectory Segmentation to separate each demonstration into trajectory segments of types (1) \textit{maintain-speed}: subtasks requiring precision or dexterity, and (2) \textit{speed-up}: simpler motions that a robot can exploit to complete a task in less time than the human demonstrator. 

The VLM's input $(V, p)$ consists of an entire video demonstration $V = \{y_0, y_1, ..., y_T\}$ along with a text prompt $p$ describing the trajectory segmentation objective with general task-invariant guidelines. We provide a few in-context examples of diverse tasks (excluding the target task) to enhance reasoning and encode optimal strategies.
The model outputs an ordered set of segments $\{(\tau_0, l_0), ..., (\tau_{k}, l_{k})\}$, where $\tau$ denotes the start and end index for the segment, and $l$ is the label. 

We specifically leverage Qwen3-VL-32B-Instruct-FP8~\cite{Qwen3-VL} to autonomously generate the labels for the entire dataset. Details of the VLM implementation, including prompts and sampling hyperparameter are detailed in Appendix~\ref{appx:vlm_implementation}.
We evaluate the computational cost for segmenting for all the tasks shown in \fig{experiments_tasks} using a single H200 GPU. We observe that VOLT generates the labels for all $250$ demonstrations in our dataset in $\sim32$ minutes with batch inference, where the longest demonstrations average $1000$ frames. To further assess VOLT’s applicability to large-scale datasets and longer horizon tasks, we label the 10 longest demonstrations from DROID~\cite{khazatsky2024droid}, where each selected demonstration contains over $3500$ frames: in this case, VOLT requires $12.5$ minutes to segment all demos. We note that for demonstrations containing $3000$ frames is best to either downsample the video or separate it into chunks for parallel segmentation over shorter context windows.

During training, the robot now has access to a dataset $\mathcal{D}$ with samples $\{(x,y, l)\}$ and user's chosen downsampling factor $n$. The robot then trains its policy on the resampled dataset.
We apply downsampling only for observations $o_t$ when label $l_t =$ \textit{speed-up}. 
To directly compare against the previous approaches, we use downsampling factors $n = 2$ and $n =4$. 
Our results compare VOLT to the alternatives in \fig{hypotheses_test}.
For the pick and place task we find that models trained with VOLT maintain or achieve a higher success rate than \textit{Normal Speed}, while for the cup pushing task there is a slight decrease in performance, usually due to errors in gripper placement behind the cup. 
As expected, only speeding up in the segments identified by VOLT yield polices that are slower than naively downsampling the entire trajectory.
But this approach ensures the robot completes the task faster than the \textit{Normal Speed} policy without sacrificing its success rate.

\subsection{What Limits VOLT's Acceleration?} \label{sec:method_d}

Our previous experiments show that we can leverage VOLT to accurately estimate when to accelerate. But what limits the downsampling factor $n$? Here we discuss the practical factors that prevent designers from achieving even faster motions.

\p{Policy Inference Delay} 
The robot's high-level policy $\pi$ converts its observations into actions.
To prevent unintended pauses during evaluation, these actions must be readily available to the robot.
However, when using diffusion policies the inference process --- i.e., reasoning over images and generating the action sequence --- is computationally expensive and typically exceeds the high-level control frequency. 
We mitigated this challenge by performing policy inference and action execution asynchronously. 
But there still remains a fundamental restriction on the control frequency $f$ as a function of the inference delay $t_{\text{inference}} < \frac{H}{f}$. 
Although increasing horizon $H$ allows us to also increase $f$, executing longer action horizons prevents the robot from reacting to changes in the environment.

\p{Open-Loop vs. Closed-Loop}
When the robot's pacing matches the human demonstrator, it can consistently update its actions in response to errors, and make corrections to improve its performance.
But if the robot maintains the same control frequency --- and accelerates its motions --- then the robot may have large errors before it queries its policy for a correction.
Taken to an extreme, if we increase $n$ such that each segment is just the start and goal waypoints, then the robot loses its ability to reason over any changes in the environment.
Designers must tune $n$ to trade-off between faster, open-loop execution and slower, closed-loop corrections.

\begin{figure*}[h!]
\centering
    \includegraphics[width=\textwidth]{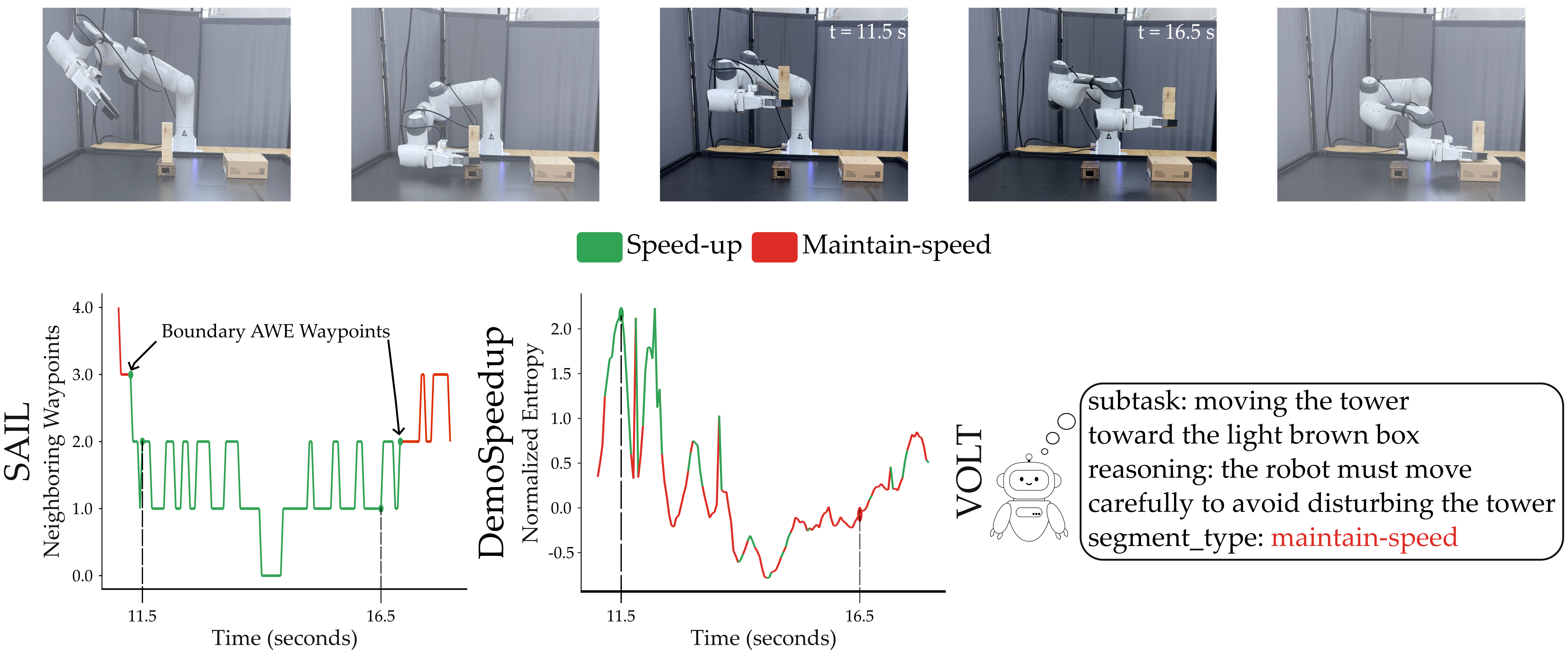}
    \caption{Example segments for transporting the block tower in the Tower Transfer task. Ideally, the robot should maintain speed during the segment from $11.5$s -- $16.5$s, since speeding up causes the blocks to fall over.
    \textit{SAIL} labels the entire sequence as \textit{speed-up} because a straight-line motion has low complexity. \textit{DemoSpeedup} oscillates between modes while focusing on the entropy in the human's demonstrations. Both of these approaches cause the tower to fall apart during transfer. \textit{VOLT} correctly recognizes that speeding up could lead to failure, and accordingly labels the segment as \textit{maintain-speed}.}
    \label{fig:labels}
\end{figure*}

\p{Experiment Results} 
To test these concepts we increased $n$ until the low-level controller was unable to track the reference positions in $A$. 
The results are summarized in \fig{hypotheses_test} as \textit{VOLT xLarge}.
Compared to the VOLT alternatives, \textit{VOLT xLarge} has significantly lower success rates in both tasks (decreasing performance for \textit{Pick and Place} by $29.7\%$ and \textit{Push Cup} by $29.2\%$).
Interestingly, the completion time with \textit{VOLT xLarge} is not significantly lower: this is because the robot's policy makes several mistakes due to tracking errors, and takes added time to redo or correct its motions. Example failures include missing the cup when trying to grasp, knocking the cup over by mistake, or pushing the cup at the wrong angle.
Overall, we conclude that the designer cannot simply increase $n$ until they reach their robot's physical or control limits. The time delay associated with the policy imposes constraints that stack with any delays in the sensors (e.g., cameras are often limited to $60$ frames per second).

%% file: 5_experiments.tex
\section{How Does VOLT Compare to Alternatives?} \label{sec:experiment}


In this section we compare \textit{VOLT} against imitation learning approaches designed to accelerate policy execution. 
We focus on comparing both the performance and speedup magnitude over a standard policy trained on the human's demonstrations. We utilize the same tabletop setup as in Section~\ref{sec:method} and evaluate additional short-horizon and long-horizon tasks (see \fig{experiments_tasks}).

\p{Baselines}
We compare \textit{VOLT} to the two most similar baselines: \textit{DemoSpeedup} and \textit{SAIL}.
Both of these approaches have the same general structure as \textit{VOLT}, in that they use a high-level controller $\pi$ to select the desired positions and a low-level controller to reach those positions as quickly and accurately as possible.
The differences are explained below:
\begin{enumerate}
    \item \textit{DemoSpeedup}: Learns a proxy policy on the original dataset to estimate the conditional action entropy of each demonstration~\cite{guo2025demospeedup}. The authors employ hierarchical density-based spatial clustering (HDBSCAN)~\cite{mcinnes2017hdbscan} to identify low-entropy clusters representing regions where the robot requires precise motion, while high-entropy outliers represent motions that do not require precision.
    \item \textit{SAIL}: Extracts waypoints from each demonstration using AWE~\cite{shi2023waypoint} and estimate motion complexity by performing DBSCAN~\cite{ester1996density} clustering on the selected waypoints. Similar to DemoSpeedup, SAIL identifies clustered regions as precise motion and outliers as the non-precision regions. In the original implementation these labels are predicted at runtime to modulate control frequency. For a direct comparison to our approach, we use the generated labels to downsample the dataset before training.
\end{enumerate}

\p{Tasks} In this section, we evaluate all approaches on five manipulation tasks. The expected progression for each task is shown in  \fig{experiments_tasks}. Both \textit{Pick and Place} and \textit{Push Cup}, were introduced in Section~\ref{sec:method}. Below, we describe the goal for the new manipulation tasks:


\begin{itemize}
    \item{\textit{Tower Transfer}}: The robot transfers a tower of freely stacked wooden blocks to a new platform.  
    \item{\textit{Plug Insertion}}: The robot picks a plug and inserts it into a socket that is fixed in place. We test on assembly ID $\#00186$ from the \textit{AutoMate} assembly benchmark~\cite{tang2024automate}.
    \item{\textit{Table Sorting}}: A long-horizon task with fruits and empty containers are scattered on a table. The robot's goal is to sort the fruits on a plate and the containers in a basket.
\end{itemize}

\p{Training} We collect demonstrations using the robot arm setup from Section~\ref{sec:method}.
For each short-horizon task we collect $\sim50$ demonstrations, and $100$ demonstrations for the long-horizon \textit{Table Sorting} task. 
Across demonstrations, we randomize the initial position of the manipulated objects within a pre-defined region, and keep the target object positions fixed. 
The same training data is provided to \textit{VOLT} and the baselines.
We then use each separate algorithm to determine which segments of these demonstrations to downsample.
Once the segments are assigned by \textit{VOLT}, \textit{DemoSpeedup}, and \textit{SAIL}, we then train separate models with identical network architectures and hyperparameters for every method to assess how the label affect policy speedup and performance. For \textit{Pick and Place}, \textit{Push Cup}, and \textit{Tower Transfer}, we apply a downsampling rate of $\times 2$ and $\times 4$ for \textit{maintain-speed} and \textit{speed-up} segments, respectively. While for \textit{Plug Insertion} and \textit{Table Sorting}, we apply $\times 1$ and $\times 3$ downsampling rates..

\subsection{How Accurately Does VOLT Segment Demonstrations?} \label{sec:exp_a}

The key difference between \textit{VOLT} and the baselines is how they segment the trajectory.
When comparing methods, we focused on how accurately each approach decomposes the human's demonstrations into labeled subtasks:  $l = \textit{speed-up}$ for simple segments and $l = \textit{maintain-speed}$ for precise motions. In \fig{labels} we show an example for Tower Transfer. Intuitively, the robot should move quickly to pick up the blocks, and then maintain the demonstrated speed while transferring the tower. Given the exact same demonstration, \textit{SAIL} incorrectly speeds up the entire motion of transferring the tower, while \textit{DemoSpeedup} alternates between accelerating and maintaining speed; only \textit{VOLT} correctly recognizes that the transfer motion should not be accelerated.
Note that this labeling is not possible with alternatives \cite{kim2025espada}, since they rely on distance features --- and here the relevant feature is how the block tower may collapse.

In general, while we observe that \textit{SAIL} correctly labels gripper actuation as \textit{maintain-speed} from the heuristics of AWE, this method more aggressively categorizes regions as \textit{speed-up} compared to the alternative methods.
On the other hand, \textit{DemoSpeedup} is overly conservative and labels most parts of the task as \textit{maintain-speed}; this most likely occurs because a fixed goal position for the manipulated object produces a lower entropy value around these repetitive movements. While \textit{DemoSpeedup} is more conservative, the entropy estimation often leads to unexpected switching between \textit{maintain-speed} and \textit{speed-up} for precise segments.

\begin{figure*}[h]
    \begin{center}
    \includegraphics[width=\textwidth]{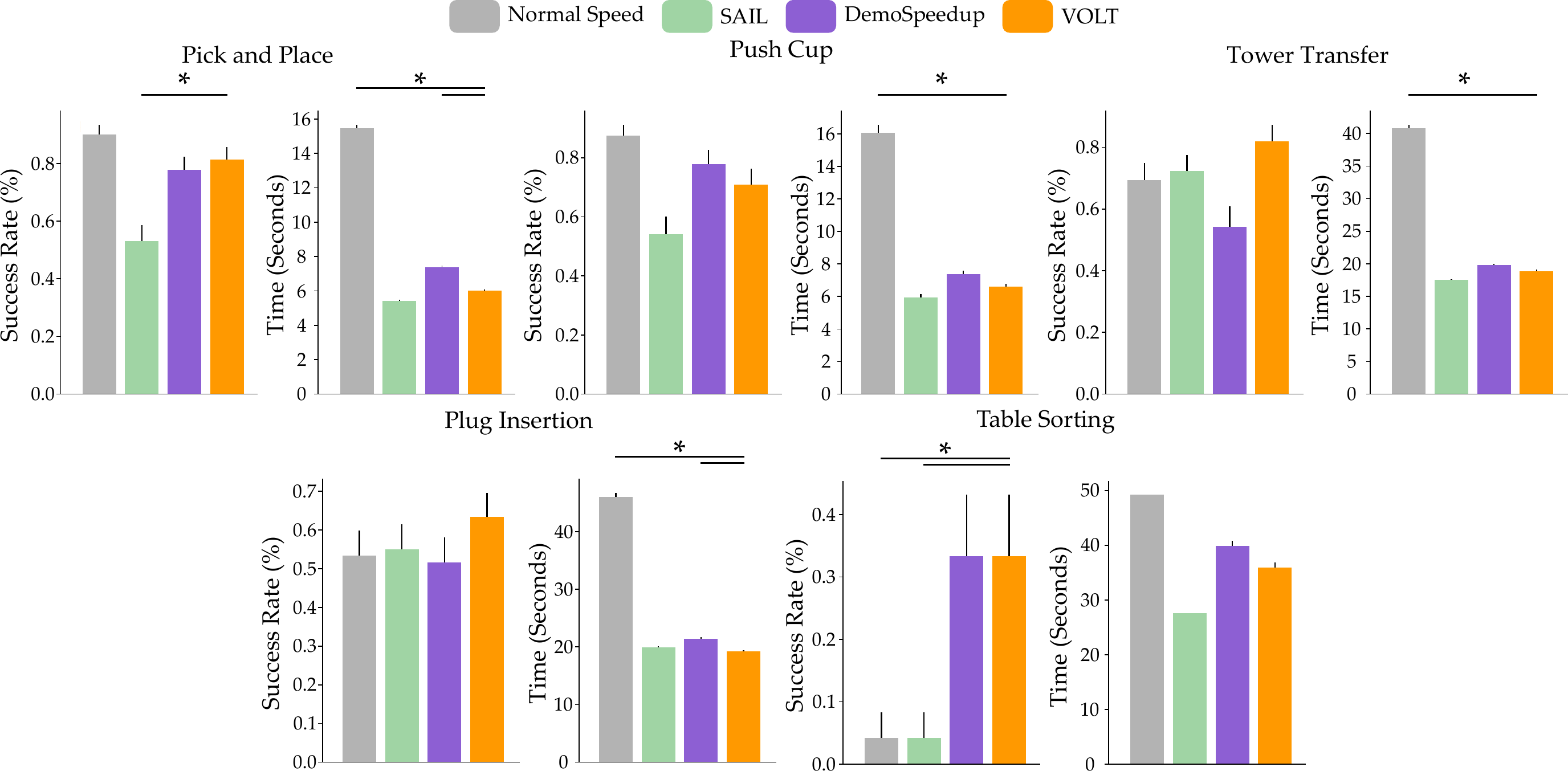}
    \caption{Experimental results for Section~\ref{sec:experiment}. We compare VOLT against state-of-the-art baselines across five manipulation tasks: \textit{Pick and Place}, \textit{Push Cup}, \textit{Tower Transfer}, \textit{Plug Insertion}, and \textit{Table Sorting} (see \fig{experiments_tasks}). All methods use the same network architecture and downsampling rate. We report results for a single model in \textit{Table Sorting} and average results over $3$ training runs for the short-horizon tasks. In our results \textit{SAIL} overestimates \textit{speed-up} segments (leading to faster but less successful policies), while \textit{DemoSpeedup} often misses \textit{speed-up} segments (leading to slower but consistent policies). We find that \textit{VOLT} balances between these extremes: significantly reducing task completion time, and achieving the second largest speedup across all tasks while maintaining similar performance to \textit{Normal Speed}. An $*$ denotes statistical significance, and error bars indicate standard error.}
    \label{fig:experiment}
    \end{center}
\end{figure*}

\subsection{Does VOLT Successfully Accelerate Policy Execution?}

Our experimental results are summarized in \fig{experiment}. A one-way ANOVA showed that policies trained with \textit{VOLT} complete tasks significantly faster than \textit{Normal Speed} across all tasks $(p < 0.001~\text{for all})$. \textit{VOLT} achieved up to x$2.57$ speedup in the \textit{Pick and Place} task, and an average $\sim$ x$2.13$ over \textit{Normal Speed} without negatively impacting performance. In particular, \textit{VOLT} significantly outperforms training with the original demonstration in Table Sorting ($p < 0.05)$, where \textit{VOLT} exhibits a more consistent strategy of handling one object at a time, while \textit{Normal Speed} often switches which object to approach while moving towards them. We also observe a performance gain on \textit{Tower Transfer} and \textit{Plug Insertion} tasks. Consistent with prior work, we attribute this increase in success rate to the reduced decision-making horizon of downsampled demonstrations~\cite{shi2023waypoint}, which reduces compounding errors. 

As compared to the baselines \textit{VOLT} consistently obtains the second largest speedup. 
\textit{SAIL} is the fastest --- because it assigns the most \textit{speed-up} segments --- but its overly aggressive choices cause the robot to make mistakes that detrimentally affect performance in Pick and Place and Push Cup. 
By contrast, \textit{DemoSpeedup} most frequently labels segments as \textit{maintain-speed}, resulting in a slower policy with similar success rates.
Overall, our results suggest that \textit{VOLT} is capable of correctly estimating when to accelerate.

\subsection{Ablation Study}
The key components in our method to enable VLMs to segment demonstrations are task-invariant guidelines, a few in-context examples, and the elicited reasoning. We ablated them on the \textit{Plug Insertion} task.

\begin{itemize}
    \item{\textit{No Examples}}: We test our method without in-context examples to test the VLM's ability to identify segments without including visual guidance.
    \item{\textit{No Examples + Simple Prompt}}: We additionally remove the task-invariant textual guidance and generated reasoning to evaluate the VLM's baseline performance.
    \item{\textit{Smaller VLM}}: We generate labels with our proposed configuration and only change the VLM to Qwen3-VL-8B-Instruct-FP8 to test how VLM size affects performance.
\end{itemize}


\begin{table}[ht]
\centering
\small
\setlength{\tabcolsep}{4pt}

\caption{Ablation study results. 
\textit{Normal Speed} policies achieve $53.33\%$ success rate and take $46.1$ seconds to complete the task.}
\label{tab:ablation}

\resizebox{\columnwidth}{!}{
\begin{tabular}{lcc}
\toprule
\textbf{Method} & \textbf{Success Rate (\%)} & \textbf{Time (s)} \\
\midrule
No Examples                 & 48.33 & 19.27 \\
No Examples + Simple Prompt & 31.67 & 18.93 \\
Smaller VLM                & 46.67 & \textbf{17.88} \\
VOLT (Ours)                & \textbf{63.33} & 19.27 \\
\bottomrule
\end{tabular}
}
\end{table}

\p{Results} We find that for \textit{No Examples}, removing the in-context examples results in two main differences in the estimation of segment boundaries for some demonstrations compared to our full configuration. Without examples, the VLM becomes more conservative for \textit{maintain-speed} segments; specifically, in the generated labels, the model classifies the alignment of the gripper and the target object to start sooner than with \textit{VOLT}. Additionally, there are cases where the model will assign more \textit{speed-up} segments mainly by deciding to speed up as soon as the gripper closes or the robot places the socket. As shown in Table~\ref{tab:ablation} this differences result in policies that are significantly faster than \textit{Normal Speed} with a slight negative impact to the success rate. In the case of only providing the model with the goal of the segmentation task \textit{No Examples + Simple Prompt}, without the use of any visual or textual guidance, we observe that, similar to the \textit{No Examples} variant, the model does recognize the correct order of events that occur in the video demonstration, and it is capable of then accordingly. However, removing all sources of guidance leads to the VLM poorly estimating the end of each segment, and typically estimating the end of each subtask earlier than its actual timing, which causes the resulting policy to speed up wrongfully during grasping. Consequently, this variant negatively impacts task performance, causing a significant drop in success rate compared to \textit{Normal Speed} policies.

The segmentation generated by the smaller VLM variant results in labels that allocate more frames to \textit{speed-up} segments and reduce the boundaries for \textit{maintain-speed} segments. Notably, this model failed to recognize the correct frame where the robot completes an interaction with an object, for example, during the grasping of the socket, or when the robot finishes inserting the socket into the plug. As compared to the larger capacity model, the smaller VLM has a more aggressive segmentation style, which results in faster policies at the cost of task success rate. Overall, we find that each design choice significantly contributes to balancing the speed-performance trade-off for efficient policies.

%% file: 6_conclusion.tex
\section{Practical Considerations}\label{sec:considerations}


We summarize the key design choices identified in this work to effectively implement faster-than-demonstration policies in practice.

\begin{itemize}

    \item \p{Visual cues are necessary} While utilizing the robot's information to segment demonstrations can be effective, throughout our experiments in Section~\ref{sec:experiment}, we demonstrate that video reasoning is crucial to identify when to \textit{maintain-speed} for simple but precise motions (e.g., transferring the tower).
    
    \item \p{Eliminate sensor-inference delays} Effective faster-than-demonstration policies require the robot to always have actions available to execute, but synchronous inference introduces unintended pauses. As described in Section~\ref{sec:method}, we instantiate diffusion policies to predict action chunks and perform asynchronous inference to eliminate these delays. In addition, we apply EAG to ensure consistency between asynchronously predicted actions. We recommend that practitioners utilize asynchronous inference when the effective policy frequency is slower than the intended high-level frequency, along with action consistency methods~\cite{arachchige2025sail,black2025training}.
    
    \item \p{Reduce the receding control horizon} As the robot operates at greater speeds, it will become less reactive. Our results from \textit{VOLT xLarge} in Section~\ref{sec:method_d} suggest that instead of executing a long sequence of open-loop actions, it is better to reduce the policy action horizon and query $\pi$ more frequently to correct potential errors. We can then pass the output actions $A$ to the low-level controller as soon as they are updated.

    \item \p{Optimizing the low-level controller} In our hierarchical paradigm, the policy sets goal positions at a slower frequency, and the low-level controller tracks them via velocity or torque commands. Tuning low-level gains for fast, accurate tracking is essential to maximize acceleration to prevent the policy from reaching out-of-distribution states.  

    \item \p{Smooth transition between fast and precise segments} Abrupt changes in acceleration cause motion jittering, which prevents the robot from reliably executing tasks near the boundaries of segments. In our experiments, we apply action interpolation near these boundaries. In practice, we recommend downsampling rates starting with $2$x-$4$x, which have been tested to perform consistently well.
\end{itemize}

\section{Conclusion} \label{sec:conclusion}

In this paper, we presented VOLT, a data augmentation framework to speed up task execution without degrading task performance. 
Our approach recognizes which segments of a task a robot can complete faster than demonstrated, and segments where the robot should operate similarly to the demonstrator. 
Specifically, we identify these segments based on the context of the task and the images of the environment by leveraging the video understanding capabilities of VLMs.
We then trained accelerated robot policies by downsampling the identified regions at different rates. 
Our experiments suggest that this approach enables robots to execute task more efficiently by increasing execution speed, while maintaining similar performance to models trained on the original dataset.

\p{Limitations and Future Work} Our results suggest that VOLT is a step towards learning policies that balance between task performance and execution speed. However, our current implementation requires downsampling rates to be manually defined and applied uniformly. Future work should automate this process, possibly guided by the VLM's own uncertainty, employing trajectory compression algorithms, or a hybrid approach combining VOLT with similar methods that have access to the low-level robot information.

%% file: 7_declaration.tex
\section{Declarations}
\label{sec:declarations}

\p{Funding}
This work was supported in part by Mitsubishi Electric Research Laboratories and the work of authors R. Ramirez Sanchez, D. Evans, and D. Losey was also funded in part by the National Science Foundation (NSF) under Grant No. \#2129201.

\p{Conflict of Interest} The authors declare that they have no conflicts of interest.

\p{Author Contribution} R.R.S.: Conceptualization, Investigation, Methodology, Data Collection, Experimentation, Writing - original draft. 
D.E.: Data Collection, Experimentation.
D.L.: Conceptualization, Supervision, Funding Acquisition, Writing - review and editing.
S.J.: Conceptualization, Supervision, Methodology, Funding Acquisition, Writing - review and editing.

%% file: 8_appendix.tex
\appendix
\onecolumn

\section{Appendix}

\subsection{Policy Training Hyperpameters}
In Table~\ref{tab:policy_hyperparams} we list the hyperparameter configuration for the diffusion policy implementation shared across our method and baselines.

{%
\captionsetup[table]{justification=centering}
\begin{table*}[ht]
\vspace{-1.5em}
\centering
\caption{Hyperparameter for diffusion transformer instantiation and training.}
\label{tab:policy_hyperparams}
\setlength{\tabcolsep}{6pt}
\begin{tabular}{|l|l|}
\hline
\textbf{Hyperparameter} & \textbf{Value} \\
\hline

\multicolumn{2}{|c|}{\textbf{Policy Architecture}} \\
\hline
Observation Horizon & 2 \\
\hline
Prediction Horizon & 32 \\
\hline
Action Horizon & 16 \\
\hline
Attention Layers & 8 \\
\hline
Attention Heads & 4 \\
\hline
Embedding Dimension & 256 \\
\hline
Embedding Dropout & 0.0 \\
\hline
Attention Dropout & 0.3 \\
\hline

\multicolumn{2}{|c|}{\textbf{Future Action Conditioning}} \\
\hline
Future Action Horizon & 4 \\
\hline
Conditioning Probability & 0.3 \\
\hline

\multicolumn{2}{|c|}{\textbf{Noise Scheduler}} \\
\hline
Algorithm & DDIM \\
\hline
Beta Schedule & squaredcos\_cap\_v2 \\
\hline
Training Iterations & 100 \\
\hline 
Inference Iterations & 10 \\
\hline

\multicolumn{2}{|c|}{\textbf{Exponential Moving Average}} \\
\hline
Inverse Gamma & 1.0 \\
\hline
Power & 0.75 \\
\hline

\multicolumn{2}{|c|}{\textbf{Training Configuration}} \\
\hline
Image Size & $\left[224, 224\right]$ \\
\hline
Batch Size & 64 \\
\hline
Epoch & 1000 \\
\hline
Learning Rate & $1e-4$ \\
\hline
Optimizer & AdamW \\
\hline
Weight Decay & $1e-6$ \\
\hline
LR Scheduler & Cosine linear warmup \\
\hline
Warmup steps & 1000 \\
\hline
\end{tabular}
\vspace{-2em}
\end{table*}
}

\subsection{VLM Implementation Details}
\label{appx:vlm_implementation}

We instantiate the VLM using vLLM~\cite{kwon2023efficient} on a single H200. The model is loaded with a context window of $262$k tokens and prefix caching for in-context examples and system prompt. In Figure~\ref{fig:system_prompt} and Figure~\ref{fig:user_prompt}, we show the system and user prompt used in our experiments. In Table~\ref{tab:vlm_hyperparams} we list the sampling hyperparameter values for reproducing our results, most parameters follow those suggested by Qwen's developers, except for the temperature and presence penalty that were tuned for better performance on our task.

{%
\captionsetup[table]{justification=centering}
\begin{table*}[h]
\vspace{-1em}
\centering
\caption{VLM sampling hyperparameters for video segmentation.}
\label{tab:vlm_hyperparams}
\setlength{\tabcolsep}{6pt}
\begin{tabular}{|l|l|}
\hline
\textbf{Hyperparameter} & \textbf{Value} \\
\hline
Random Seed        & 3047 \\
\hline
Top P              & 0.8 \\
\hline
Top K              & 20 \\
\hline
Temperature        & 0.4 \\
\hline
Repetition Penalty & 1.0 \\
\hline
Presence Penalty   & 0.5 \\
\hline
\end{tabular}
\end{table*}
}

\onecolumn
\begin{center}
\begin{tcolorbox}[breakable, colback=gray!5,colframe=black,title=System Prompt]
You are a helpful assistant that generates labels for a given video of a robot performing a task.

Your goal is to help the robot perform the task as fast as possible without causing it to fail by segmenting the video into different subtasks. You first need to analyze the video and then localize all the subtasks performed to complete the main task. For each subtask, you need to identify its start and end timestamps and classify the subtask into one of the valid segment types: speed-up, maintain-speed. Your generated response must help the robot increase its task efficiency while maintaining precise and critical movements.

\textbf{Subtask Localization and Classification Guidelines:}
\begin{itemize}
\item
The robot should speed up whenever possible, this refers to cases that do not require delicate environment interactions.
\item
Maintain-speed only when it is necessary for the robot to work as demonstrated, mainly subtasks that require dexterity or precision.
\item 
The robot is a physical system that cannot instantly change speed, you need to allocate enough frames on slower segments when transitioning between subtasks that do not share the same category to allow the robot to smoothly slow down and avoid jerky motions.
\item
Given direct grasping of an object, the robot has enough force to securely hold it regardless of speed.
\item
Fast speed will often lead to unexpected motion of objects not directly grasped by the gripper.
\item
Default the events where the robot is actively avoiding obstacles without an exaggerated motion, i.e., the robot barely clears the obstacle as precise segments.
\item
When the robot is done with the task and interacting with objects, the robot must quickly reach its final position.
\end{itemize}  

\textbf{Generation Rules:}
\begin{itemize}
\item
Provide a concise reasoning for which label is appropriate for each segment must be provided before making a decision.
\item 
The list of segments must cover the entire video.
\item 
Only leave one frame gaps between segments.
\item 
All robot alignment subtasks must be categorized as precise.
\end{itemize}

The following is a non-exhaustive list of common subtasks:
\begin{itemize}
\item
Moving towards objects
\item 
Transporting objects
\item 
Avoiding obstacles
\item 
Grasping/Placing
\item 
Insertion/Retrieval
\item 
Pushing
\item 
Pouring
\item 
Wiping
\end{itemize}

This is not a complete list of possible subtasks. If any identified event is not included in this list, be sure to describe it accordingly.
\\
The generated response must use the ``mm:ss.ff" format for all timestamps. Final output format:\\
\textasciigrave\textasciigrave\textasciigrave json
\begin{lstlisting}[language=json]
[
    {
        "start_time": "mm:ss.ff",
        "end_time": "mm:ss.ff",
        "subtask": "concise subtask description",
        "reasoning": "reasoning of which segment_type should be used",
        "segment_type": "speed-up/maintain-speed"
    },
    ...
]
\end{lstlisting}
\textasciigrave\textasciigrave\textasciigrave
\end{tcolorbox}
\captionof{figure}{System prompt for video segmentation}
\label{fig:system_prompt}
\end{center}
\FloatBarrier

\begin{center}
\begin{tcolorbox}[colback=gray!5,colframe=black,title=User Prompt]
Given the current task description: \texttt{<|video-description|>}, help the robot complete the task faster than demonstrated in the video (00:00.00 - \texttt{<|video-length|>})
\end{tcolorbox}
\captionof{figure}{User prompt for video segmentation}
\label{fig:user_prompt}
\end{center}